\documentclass[lettersize,journal]{IEEEtran}
\usepackage{amsmath,amsfonts}
\usepackage{algorithmic}
\usepackage{algorithm}
\usepackage{array}
\usepackage[caption=false,font=normalsize,labelfont=sf,textfont=sf]{subfig}
\usepackage[T1]{fontenc}
\usepackage{textcomp}
\usepackage{stfloats}
\usepackage{url}
\usepackage{verbatim}
\usepackage{graphicx}
\usepackage{cite}
\usepackage{multirow}
\usepackage{tablefootnote}
\usepackage{hyperref}
\hypersetup{
  colorlinks,
  urlcolor=blue,
  linkcolor=blue,       %%修改此处为你想要的颜色
  anchorcolor=blue,  %%修改此处为你想要的颜色
  citecolor=blue,        %%修改此处为你想要的颜色，例如修改blue为red
}
\usepackage{array}

\usepackage{tikz,xcolor,hyperref}

\definecolor{lime}{HTML}{A6CE39}
\DeclareRobustCommand{\orcidicon}{
\begin{tikzpicture}
\draw[lime, fill=lime] (0,0)
circle[radius=0.16]
node[white]{{\fontfamily{qag}\selectfont \tiny \.{I}D}}; 
\end{tikzpicture}
\hspace{-2mm}
}
\foreach \x in {A, ..., Z}{%
\expandafter\xdef\csname orcid\x\endcsname{\noexpand\href{https://orcid.org/\csname orcidauthor\x\endcsname}{\noexpand\orcidicon}}
}

\begin{document}

\title{Block shuffling learning for Deepfake Detection}

\author{Sitong Liu\hspace{-1.5mm}\orcidC{}, Zhichao Lian\hspace{-1.5mm}\orcidA{},~\IEEEmembership{Member,~IEEE}, Siqi Gu\hspace{-1.5mm}\orcidD{}, Liang Xiao\hspace{-1.5mm}\orcidB{},~\IEEEmembership{Member,~IEEE}
        % <-this % stops a space
  \thanks{
    Manuscript received September 20, 2022. This work was supported in part by the National Natural Science Foundation of China under Grant 61871226, Grant 61571230, Grant 61802190, and Grant 61906093, in part by the Jiangsu Provincial Social Developing Project under Grant BE2018727, in part by the Open Research Fund in 2021 of Jiangsu Key Laboratory of Spectral Imaging \& Intelligent Sense under Grant JSGP202101, and Grant JSGP202204, in part by the National Key R\&D Program of China under Grant 2021YFF0602104-2.
  }
}

\markboth{IEEE TRANSACTIONS ON BIOMETRICS, BEHAVIOR, AND IDENTITY SCIENCE}
{Block shuffling learning for Deepfake Detection}


\maketitle

\begin{abstract}

Deepfake detection methods based on convolutional neural networks (CNN) have demonstrated high accuracy. \textcolor{black}{However, these methods often suffer from decreased performance when faced with unknown forgery methods and common transformations such as resizing and blurring, resulting in deviations between training and testing domains.} This phenomenon, known as overfitting, poses a significant challenge. To address this issue, we propose a novel block shuffling regularization method. Firstly, our approach involves dividing the images into blocks and applying both intra-block and inter-block shuffling techniques. This process indirectly achieves weight-sharing across different dimensions. Secondly, we introduce an adversarial loss algorithm to mitigate the overfitting problem induced by the shuffling noise. Finally, we restore the spatial layout of the blocks to capture the semantic associations among them. \textcolor{black}{Extensive experiments validate the effectiveness of our proposed method, which surpasses existing approaches in forgery face detection. Notably, our method exhibits excellent generalization capabilities, demonstrating robustness against cross-dataset evaluations and common image transformations.} Especially our method can be easily integrated with various CNN models. Source code is available at \href{https://github.com/NoWindButRain/BlockShuffleLearning}{Github}.

% Although deepfake detection methods based on convolutional neural network (CNN) achieve good accuracy, results illustrate that the performance degrades significantly when images undergo some common transformations (like resizing and blurring) which leads to deviations between training and testing domains. We identify this as the overfitting problem and propose a novel block shuffling regularization method to solve it. Firstly, we divide the images into blocks and perform intra-block and inter-block shuffling to achieve weight-sharing across different dimensions indirectly. Secondly, we propose an adversarial loss algorithm to overcome the overfitting problem brought by the shuffling noise. Finally, we restore the spatial layout of the blocks to model the semantic associations among them. \textcolor{black}{Extensive experiments show that our proposed method achieves state-of-the-art performance in forgery face detection, including good generalization ability against cross-dataset and common image transformations.} Especially our method can be easily integrated with various CNN models. Source code is available at \href{https://github.com/NoWindButRain/BlockShuffleLearning}{Github}.
\end{abstract}


\section{Introduction}

\begin{figure}[htp]
  \centering
  \includegraphics[width=\linewidth]{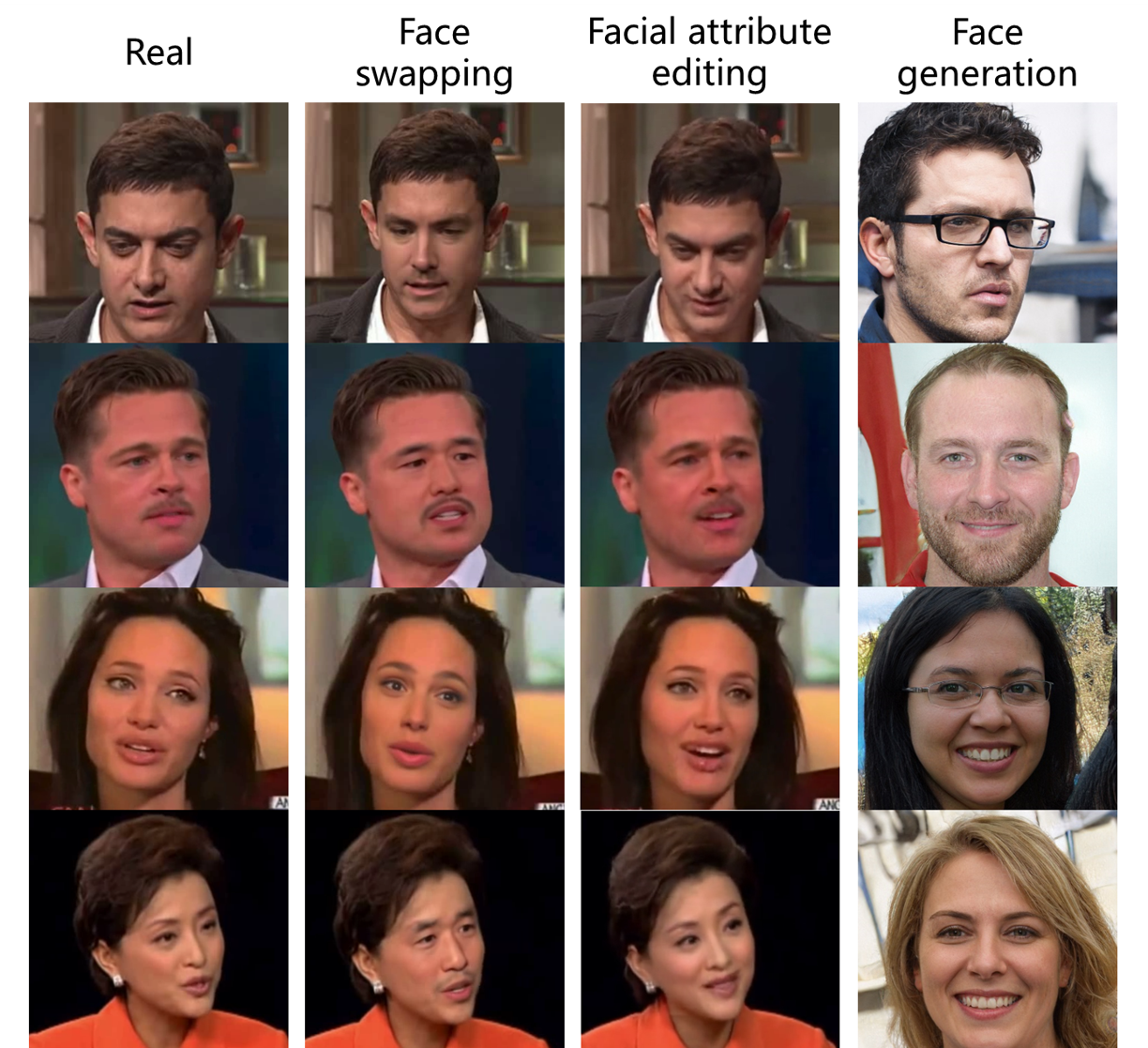}
  \caption{An illustration of various real and fake faces. Both real faces and swapped faces are taken from CelebDF(v2)~\cite{Li2020CelebDFAL}, the faces edited by the expression adopt GANimation~\cite{Pumarola2018GANimationAF}, and the generated faces are taken from StyleGAN~\cite{stylegan}.}
  \label{fig:realfake}
\end{figure}

\IEEEPARstart{I}{n} recent years, the field of deep learning has witnessed significant advancements, particularly in the area of face forgery techniques. Notably, these techniques, which rely on deep learning, have demonstrated substantial progress. As reported by Sensity in a comprehensive survey\cite{sensity}, the proliferation of harmful deepfake videos crafted by skilled creators on the Internet has been alarmingly escalating, with the number of such videos approximately doubling every 6 months. These face forgery techniques can be categorized based on their intended targets, namely face swapping\cite{Nirkin2019FSGANSA}, facial attribute editing\cite{Siarohin2019FirstOM, Choi2018StarGANUG}, and face generation\cite{Karras2019ASG}, as depicted in Fig. \ref{fig:realfake}. Of particular influence are face swapping methods, with DeepFakes\cite{DeepFakes} serving as a prominent example. These techniques possess the capability to manipulate the facial identity information within an image. Consequently, the widespread dissemination of fabricated images on the Internet has engendered significant security risks, thus necessitating the advancement of deepfake detection technologies. Presently, the majority of detection algorithms\cite{Du2020TowardsGD} approach the problem as a classification task, leveraging data-driven training of convolutional neural networks (CNNs) to identify fake faces and achieve good performance. \textcolor{black}{However, the continuous update of forgery methods and the propagation of images often leads to varying degrees of degradation, including resizing and blurring, thereby challenging the generalization capability of detection methods.} In severe cases, deep models may exhibit satisfactory performance on the training data but encounter a considerable decline in accuracy when confronted with new data. An essential contributing factor to this phenomenon is the overfitting problem inherent in the detection model's local context.

% \IEEEPARstart{I}{n} recent years, with the rapid development of deep learning, face forgery techniques based on deep learning have made significant progress. According to a survey\cite{sensity} conducted by Sensity, the number of harmful deepfake videos crafted by expert creators on the Internet doubles roughly every 6 months. According to different forgery targets, scholars classify them as face swapping\cite{Nirkin2019FSGANSA}, facial attribute editing\cite{Siarohin2019FirstOM, Choi2018StarGANUG} and face generation\cite{Karras2019ASG} as shown in Fig. \ref{fig:realfake}. The most influential are face swapping methods represented by DeepFakes\cite{DeepFakes}, which can change the face identity information in the image. Therefore, the mass dissemination of fake images on the Internet creates security risks, promoting deepfake detection technology development. Currently, most detection algorithms\cite{Du2020TowardsGD} treat it as a classification problem, using data-driven training of CNN to detect fake faces and achieve good performance. However, images are often degraded to varying degrees during propagation (like resizing and blurring), which challenges the generalization ability of detection methods. In severe cases, deep models may perform well on training data but have a significant drop in accuracy when predicting new data. An important reason is the overfitting problem of the detection model in the local area. 

The issue of overfitting significantly affects the task of forgery detection, as the decision-making process for identifying fake images primarily relies on localized regions. Face forgery entails altering specific attributes, such as identity and expression, of a face in an image to resemble those of another face. The forgery techniques aim to bridge the visual and distribution gap between the two images. Various forgery methods introduce distinct artifacts-based features in different facial regions\cite{Rana2022DeepfakeDA}. Deepfake detection methods\cite{Chollet2017XceptionDL, Zhao2021MultiattentionalDD,Li2020FaceXF} heavily rely on the accurate perception of these local regions to identify subtle cues, such as mismatched colors, distorted shapes, and incorrect textures, caused by imperfect facial manipulation techniques. \textcolor{black}{However, these traces of forgery are inherently unstable, primarily due to differences in forgery methods and image scaling, compression, and cropping during their dissemination across the Internet.} Consequently, the relatively small and unstable local regions, combined with the vast number of parameters within the network, give rise to the overfitting problem, thus impeding the model's generalization ability. Motivated by the effectiveness of regularization techniques in addressing overfitting, we propose a novel regularization method that specifically focuses on the local regions, aiming to enhance the forgery detection performance of the model.

% The overfitting problem significantly impacts the forgery detection task because the basis for judging whether an image is fake or not is mainly concentrated in local areas. Face forgery refers to modifying specific attributes (like identity and expression) of a face in an image into the corresponding attributes of another face. The forgery method aims to bridge the visual and distribution gap between the two images. Different forgery methods leave special artefacts-based features in different areas of face images\cite{Rana2022DeepfakeDA}. Deepfake detection methods\cite{Chollet2017XceptionDL, Zhao2021MultiattentionalDD,Li2020FaceXF} mostly require a good perception of local regions to recognize small traces (like mismatched colours, distorted shapes and wrong textures) caused by imperfect facial forgery methods. However, these traces are unstable because images are often scaled, compressed, and cropped as they spread on the Internet. As a result, the relatively small unstable local regions and the large number of parameters in the network may lead to the overfitting problem and limit the model's generalisation ability. Inspired by regularization methods in solving the overfitting problem, we design a regularization method to focus on local regions to effectively improve the forgery detection performance of the model.

\IEEEpubidadjcol

This paper introduces a weight-sharing regularization-based framework for deepfake detection, referred to as Block Shuffling Learning (BSL). The proposed framework focuses on positional transformations of image pixels, maintaining their original values. It comprises two main operations: intra-block shuffling and inter-block shuffling, which facilitate indirect weight-sharing across different dimensions. In the intra-block shuffling operation, local representations are enhanced through the introduction of positional transformations. Additionally, an adversarial loss is designed to filter out noisy patterns resulting from intra-block shuffling. In the inter-block shuffling operation, the global structure is disrupted to encourage feature extraction from local regions. The semantic correlations among blocks are then modeled by restoring the original regional arrangement. It is important to emphasize that our framework represents more than a simple integration or modification of existing methods. It fundamentally differs from existing approaches in terms of usage scenarios, compositional principles, and operational logic. To the best of our knowledge, we are the first to address the issue of overfitting caused by local regions in the context of deepfake detection and propose regularization as a solution. Existing scramble-based regularization methods~\cite{KangDZY17PatchShuffle, Shen2017PatchRA} are not specifically designed for deepfake detection tasks and fail to account for the noise introduced by random shuffling operations, rendering them relatively ineffective.

% This paper proposes a weight-sharing regularization-based deepfake detection framework named Block Shuffling Learning (BSL). Our framework mainly performs a positional transformation on the pixels of the image without changing the pixel value and consists of an intra-block shuffling operation and an inter-block shuffling operation, which indirectly achieves weight-sharing in different dimensions. In the former, we enrich the local representation by intra-block shuffling and design an adversarial loss to filter noisy patterns introduced by intra-block shuffling. In the latter, we destroy the global structure by inter-block shuffling to encourage the network to extract features from local regions and model the semantic correlations among blocks by restoring the original regional arrangement. It should be emphasized that this framework is more than a simple integration and modification of existing methods. Our proposed method differs from existing methods in usage scenarios, principle composition and operation logic. To the best of our knowledge, we are the first to propose the problem of overfitting caused by local regions in the deepfake detection task and introduce regularization to solve it. Existing scramble-based regularization methods~\cite{KangDZY17PatchShuffle, Shen2017PatchRA} are not proposed for the deepfake detection task. And these methods do not consider the noise introduced by random shuffling operations. So they are relatively ineffective.

Our work makes the following main contributions:

\begin{itemize}
\item{We propose an efficient weight-sharing regularization method to address the overfitting problem in local regions, which consists of an intra-block shuffling and an inter-block shuffling. Our method can be easily adopted in a variety of CNN models.}
\item{We propose an adversarial loss to overcome the shuffling noise and model the semantic relevance of local features through the position restoration of the blocks.}
\item{\color{black}Our method has undergone extensive experimentation, which substantiates its superiority in terms of detection performance across various datasets. Consistently surpassing existing methods, our approach stands out, particularly excelling in cross-dataset and cross-image quality experiments. These results serve as compelling evidence that our regularization method effectively ensures the network's generalization capabilities in the context of forgery detection.}
% \item{\color{black}Extensive experiments demonstrate that our method achieves state-of-the-art detection performance on different datasets and consistently outperforms existing methods. In particular, our method achieves the best results in cross-dataset and cross image quality experiments, which verifies that the regularization method enables the network to maintain good generalization in the forgery detection task.}
\end{itemize}

\section{Related works}

In this section, we briefly review the current representative face forgery detection and regularisation methods relevant to our work.

\textbf{Deepfake Detection.} To cope with the security problems caused by Deepfakes, people propose deepfake detection methods from multiple insights. Early work focused on designing handcrafted features~\cite{Fridrich2012RichMF,Chierchia2011PRNUbasedDO,Chen2018FocusMD,Luks2006DetectingDI}. The mainstream deep learning-based methods focus on how to extract features from the spatial domain~\cite{Zhou2017TwoStreamNN, Afchar2018MesoNetAC, Nguyen2019CapsuleforensicsUC, Li2019ExposingDV}and frequency domain information~\cite{Durall2019UnmaskingDW, Luo2021GeneralizingFF, Qian2020ThinkingIF}. Nguyen et al.~\cite{Nguyen2019CapsuleforensicsUC} propose a detection method based on the capsule network~\cite{Sabour2017DynamicRB}. Li et al.~\cite{Li2019ExposingDV} propose that the artefacts in affine face warping as the distinctive feature to forgery detect, and achieve state-of-the-art performance based on SSPNet~\cite{Hong2021SSPNetSS}. Li et al.~\cite{Li2020FaceXF} propose the Face X-ray that focuses on the blending step of forgery and achieves state-of-the-art transferability performance. However, it cannot be used for fully synthesized images, and the performance on low-resolution images drops sharply. Durallet et al.~\cite{Durall2019UnmaskingDW} first propose using Discrete Fourier Transform (DFT) to mine abnormal information in forged face images. Luo et al.~\cite{Luo2021GeneralizingFF} propose utilising the high-frequency noises for face forgery detection. Qian et al.~\cite{Qian2020ThinkingIF} use DFT to extract frequency-aware image decomposition and local frequency statistics to improve the forgery detection performance on high-compression videos. However, the cross-data set performance dropped significantly.

% Most of the above methods define the forgery detection task as a general two classification problem, so the performance is sensitive to the quality of the data set or the data distribution. 
Since the differences between real and fake images are primarily subtle and local, researchers currently find solutions from local areas. Du et al.~\cite{Du2020TowardsGD} focus on local spatial features and learn the internal representation of local forged regions through auto-encoders to bridge the generalization differences. Zhao et al.~\cite{Zhao2021MultiattentionalDD} propose a multi-attentional network to focus on different local parts and subtle artefacts. Wang et al.~\cite{Wang2021RepresentativeFM} propose an attention-based data enhancement framework, which encourages the network to mine diverse features by occluding sensitive areas. Wang et al.~\cite{Wang2021M2TRMM} propose a multi-modal multi-scale transformer to capture the subtle manipulation artefacts at different scales. We focus on local regions to solve the forgery detection problem. The difference is that we pay more attention to the overfitting problem of the network.

\textbf{Regularization.} Generalization is an essential topic for forgery detection algorithms that rely on CNN. Regularization is an effective method to improve generalization and reduce the impact of overfitting in the training of CNN models. Two kinds of regularisation are common today: model ensemble and weight-sharing. Model ensemble refers to multiple separately trained models voting to decide the output. The voting procedure is robust to prediction errors made by individual classifiers. Using a model ensemble requires a huge computational cost, so many methods implement it implicitly. Dropout\cite{Srivastava2014DropoutAS} achieves the average of architectures by temporarily dropping the neurons in the network with a certain probability during training. Stochastic Pooling\cite{Zeiler2013StochasticPF} randomly selects activation from a multinomial distribution during training to average the hidden layer inputs. In another way, weight-sharing requires a set of weights in the network to be equal to achieve invariant transformation properties\cite{Nowlan1992SimplifyingNN} and gains generalizability to permutations of the input\cite{Ravanbakhsh2017DeepLW}. PatchShuffle\cite{KangDZY17PatchShuffle} randomly shuffles the pixels within each local patch while maintaining nearly the same global structures as the original ones. It yields rich local variations for the training of CNN. Shen et al.~\cite{Shen2017PatchRA} propose dividing the feature maps into non-overlapping patches and reordering the patches to enhance the rotation and translation invariance. It is widely used in neural architecture search (NAS) because it can improve the search efficiency and reduce instability\cite{Xie2022WeightSharingNA}. 

Most of the latest CNN models\cite{He2016DeepRL, Chollet2017XceptionDL,Tan2019EfficientNetRM} integrate multiple regularization methods. However, these methods cannot efficiently solve the overfitting problem of forgery detection tasks because most of them do not pay attention to the local area. Their works~\cite{KangDZY17PatchShuffle, Shen2017PatchRA} are similar to our proposed method. We all design shuffling methods that act on blocks. However, compared with PatchShuffle\cite{KangDZY17PatchShuffle}, we randomly select blocks for intra-block shuffling, not all blocks. And we notice the noise patterns introduced by intra-block shuffling and design an adversarial loss to solve it. PatchReordering~\cite{Shen2017PatchRA} reorder blocks according to specific rules while we shuffle randomly. And we propose a position restoration method to build correlations between blocks.

% Simplifying neural networks by soft weight-sharing
% Deep learning with sets and point clouds
% Weight-Sharing Neural Architecture Search: A Battle to Shrink the Optimization Gap 

\section{Proposed Method}

\begin{figure*}[htb]
  \centering
  \includegraphics[width=1.0\linewidth]{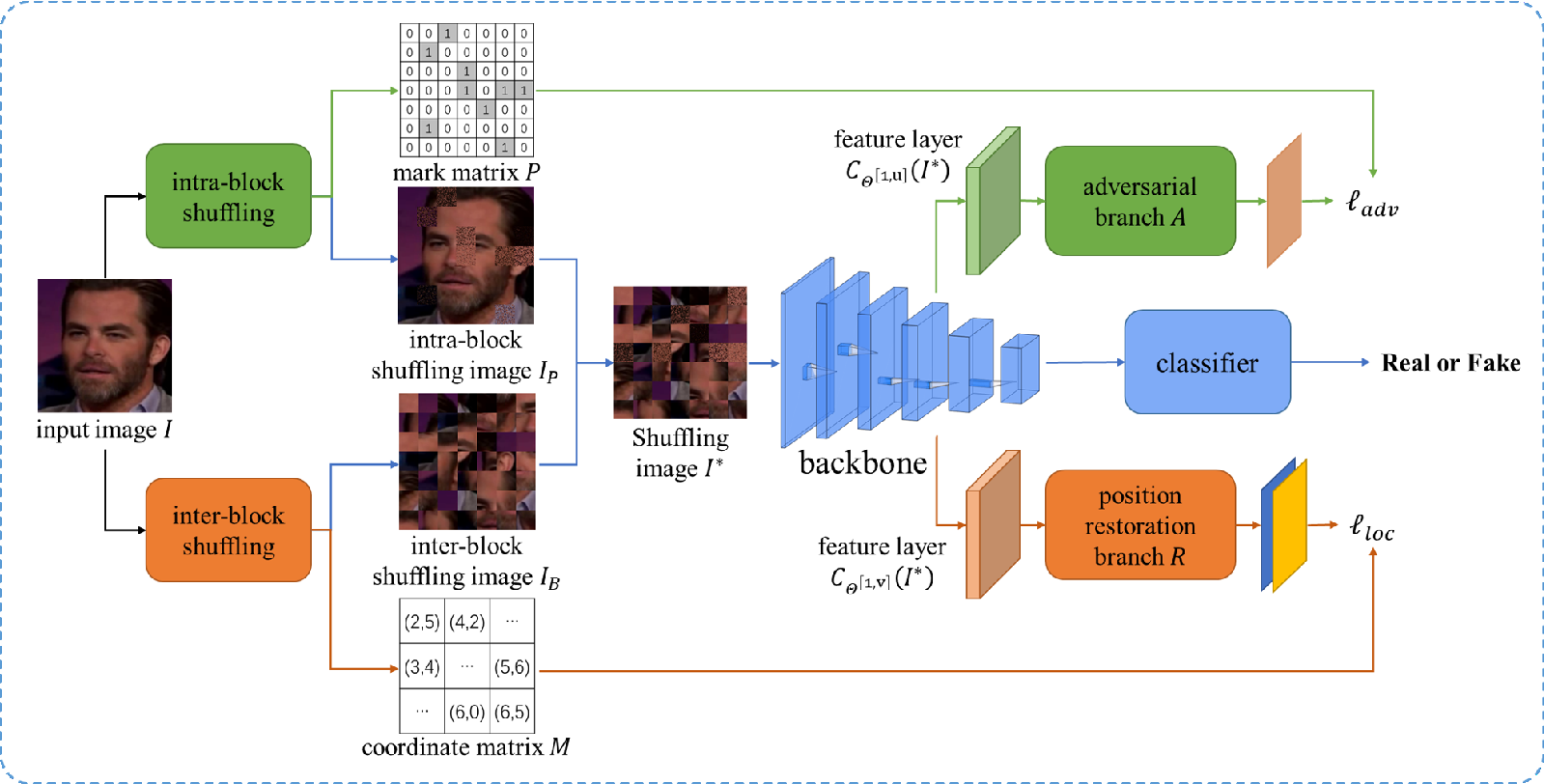}
  \caption{The framework of our method. Our method consists of four important components: 1) an intra-block shuffling module for rich local representations; 2) an inter-block shuffling module that emphasizes local features; 3) an adversarial loss module for overcoming the shuffling noise; 4) a position restoration module for modeling the semantic relevance of local features.}
  \label{fig:frame}
\end{figure*}

In this section, we commence by providing an overview of the deepfake detection problem, highlighting key observations that guide the development of our approach. Subsequently, we present a Block Shuffling Learning (BSL) framework, comprising four integral modules: an intra-block shuffling module, an inter-block shuffling module, an adversarial loss module, and an position restoration module. An illustration of the framework can be found in Fig. \ref{fig:frame}. In Sec. \ref{sec:intra}, we delve into the specifics of the intra-block shuffling module and adversarial module, while Sec. \ref{sec:inter} details the inter-block shuffling module and position restoration module.

% In this section, we first introduce observations on the deepfake detection problem. After that, we present a BSL framework, which consists of an intra-block shuffling module, an inter-block shuffling module, an adversarial module and a position restoration module. Fig. \ref{fig:frame} gives an overview of our framework. The details of four modules are introduced in Sec. \ref{sec:intra} and Sec. \ref{sec:inter} respectively. 

\subsection{Motivation}
\label{sec:overview}

\begin{figure}[htp]
  \centering
  \includegraphics[width=\linewidth]{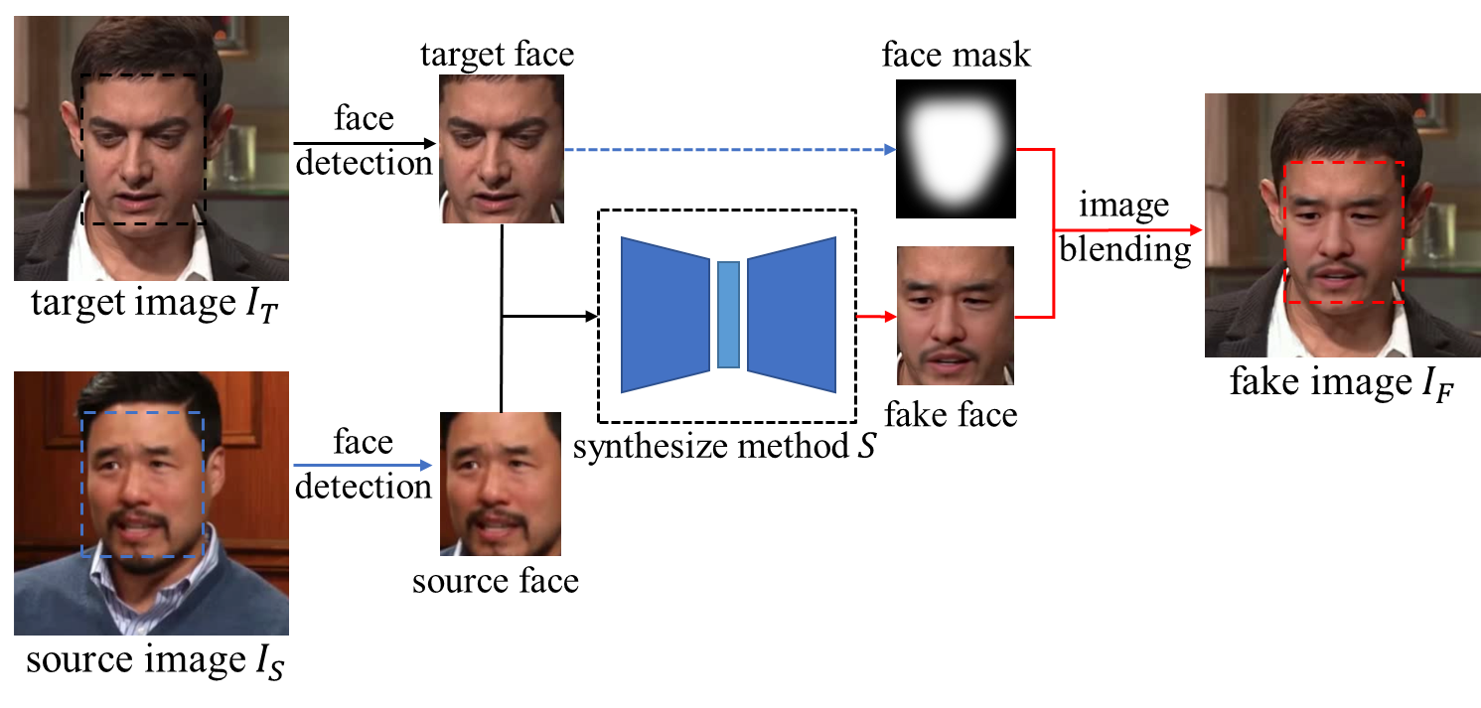}
  \caption{Process of typical face replacement.}
  \label{fig:pro}
\end{figure}

The deepfake algorithms for face swapping can be roughly divided into three categories: methods based on graphics~\cite{FaceSwap}, methods based on autoencoder~\cite{DeepFakes} and methods based on generative adversarial network (GAN)~\cite{FaceswapGAN}. These techniques only require a few face images to forge realistic fake images in a short time, which is pretty deceptive. It is difficult for even human eyes to distinguish whether the images have been manipulated~\cite{exist}. As shown in Fig. \ref{fig:pro}, face forgery technology mainly consists of three s.png to manipulate the target face and create a fake image: 1) face detection; 2) face synthesis; 3) image blending. It is noted that a fake image $I_F$ is obtained by combining two images $I_T$ and $I_S$:
\begin{equation}
  I_F = F\left(I_T, S\left(I_T, I_S\right)\right)
  \label{eq:eq11}
\end{equation}
where $I_T$ is the image of the replaced face identity and $I_S$ is the face image used for replacement. $S$ is a method of synthesizing two faces, such as making the expression in $I_S$ similar to that in $I_T$. $F$ is the method of blending the synthesized face into $I_T$, which usually requires some graphics and colour transformations. 

Face forgery detection methods generally consider the traces produced by the second and third stages in local areas of fake images~\cite{Li2020FaceXF, Zhao2021MultiattentionalDD}. However, these traces are small, varying and not easily captured by CNN. Therefore, we propose a regularization method to help the network extract local features more effectively in deepfake detection tasks. For the characteristics of focusing on local areas, we divide the image into blocks of the same size and perform random intra-block and inter-block shuffling respectively. 

% face x-ray
% multi-att

The utilization of intra-block shuffling introduces rich local variations within the image. Firstly, each pixel instance within a specific image position can be regarded as being uniformly sampled from the corresponding image block. Consequently, weight-sharing is achieved within each block across multiple iterations. Secondly, the random shuffling technique effectively applies perceptual encryption to the affected regions, prompting the network to extract distributional features while discouraging the extraction of non-distributed features.  This mechanism improves the generalization of the network by simulating the occlusion of local areas\cite{Zhong2020RandomED, Wang2021RepresentativeFM}. Lastly, local changes induced by intra-block shuffling do not adversely impact the global information of the image.  The distinct representations captured within the local and global contexts allow intra-block shuffling to play a beneficial role in the forgery detection task.

% By intra-block shuffling, the image possesses rich local variations. Firstly, the pixel instance at a specific image position can be considered to be sampled with equal probability from the image block. This enables weight-sharing within each block over multiple iterations. Secondly, the random shuffling method has the effect of perceptual encryption on the affected regions, which encourages the network to extract distributional features and prevents the network from extracting non-distributed features. This improves the generalization of the network because it has a similar effect as occluding local areas\cite{Zhong2020RandomED, Wang2021RepresentativeFM}. Finally, local changes will not affect the global information of the image. Different representations in local and global make intra-block shuffling play a positive role in forgery detection. 

Through the implementation of inter-block shuffling, the global structure of the image is intentionally disrupted. The reordering of image patches enhances the network's translation invariance capabilities\cite{Shen2017PatchRA}, thereby preventing the forgery detection task from devolving into a mere face recognition task. Each block instance can be viewed as being sampled uniformly from the blocks into which the image is divided. Given that specific regions of the face image are more likely to exhibit forgery traces, we have devised a position restoration method to establish correlations between the blocks.

% By inter-block shuffling, the global structure of the image is destroyed. The reordering of image patches improves the network's translation invariance capabilities\cite{Shen2017PatchRA} and can prevent the forgery detection task from degenerating into a face recognition task. The block instance can be viewed as being sampled with equal probability from the blocks into which the image is divided. Considering that some forgery traces are likely to appear in specific regions of the face image, we design a position restoration method to build the correlation between the blocks. 

% Patch Reordering: A Novel Way to Achieve Rotation and Translation Invariance in Convolutional Neural Networks

\subsection{Intra-block shuffling}
\label{sec:intra}

Detecting one or several specific defects in a fake image can often be an easy task. However, the challenge arises when deepfake detection methods encounter scenarios where training and test data are generated using different deepfake techniques. In such cases, the performance of these methods tends to significantly deteriorate with the addition of simple transformations. To address this limitation and enhance the generalization capabilities of the network, we propose a novel approach involving the shuffling of the internal structure of image blocks. By shuffling the internal structure of image blocks, we aim to enable the network to learn more generalized features that can effectively discriminate between genuine and fake images, even when subjected to various transformations. However, it is important to note that the process of intra-block shuffling introduces noise patterns that are potentially detrimental to the detection performance. Consequently, we propose the utilization of an adversarial loss to identify and eliminate patterns unrelated to deepfake detection, thereby enhancing the overall effectiveness of our approach.

% It is simple to detect one or several specific defects in a fake image. Though, deepfake detection methods can quickly obtain near-perfect experimental results when training data and test data are generated by the same deepfake method. Just adding some simple transformations can result in a significant drop in detection performance. We propose shuffling the internal structure of image blocks to learn more general features, thereby improving the generalization of the network. Meanwhile, intra-block shuffling introduces noise patterns that are harmful to detection. Thus, we propose an adversarial loss to reject patterns unrelated to deepfake detection.

\subsubsection{Pixel shuffling transformation}
\label{sec:ps}

Intra-block shuffling methods serve to enhance the local representations of the image while facilitating weight-sharing within each block. By applying perceptual image encryption~\cite{Chuman2019EncryptionThenCompressionSU,Sirichotedumrong2019GrayscalebasedBS,Sirichotedumrong2019PixelBasedIE} through intra-block shuffling, the network faces challenges in extracting non-distributed features from the shuffled local regions. This observation highlights a similar role of intra-block shuffling to occluding local areas~\cite{Zhong2020RandomED}, thereby promoting the network's generalization capabilities. Importantly, our method ensures that the local distribution of the image remains unchanged, thereby encouraging the neural network to extract both local statistical features and other structure-independent features. In order to achieve this objective, we have devised a transformation method, which is outlined as follows.

% Intra-block shuffling methods enrich local representations of the image and indirectly achieve weight-sharing within each block. In perceptual image encryption~\cite{Chuman2019EncryptionThenCompressionSU,Sirichotedumrong2019GrayscalebasedBS,Sirichotedumrong2019PixelBasedIE}, it is difficult for the network to extract non-distributed features from the shuffled local regions of the image. This shows that the intra-block shuffling plays a similar role as occluding local areas~\cite{Zhong2020RandomED} to improve the generalization of the network. Moreover, our method does not change the local distribution of the image and encourages the neural network to extract local statistical features and other local structure-independent features. Therefore, we designed a transformation method to achieve the goal as follows.

\begin{figure}[htb]
  \centering
  \includegraphics[width=\linewidth]{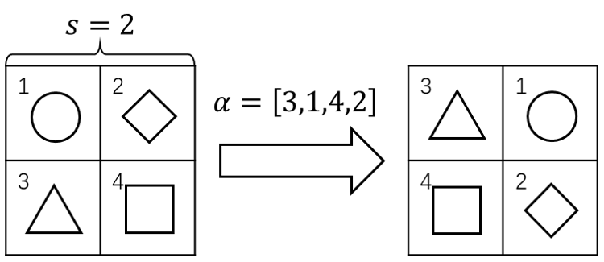}

  \caption{Process of the pixel shuffling transformation.}
  \label{fig:props}
\end{figure}

Specifically, given an input image I with a dimension of $x\times y$, we first divide the image into blocks with a size of $s\times s$, where $x$ and $y$ should be divisible by $s$. Thus, we get $m\times n$ blocks denoted by $B_{i,j}$, where $m$ and $n$ are the numbers of horizontal and vertical blocks, $i$ and $j$ are the horizontal and vertical indices, respectively. Aprilpyone et al.~\cite{Aprilpyone2020EncryptionIA} demonstrate the ability of the neural network to learn specific shuffling rules. Thus, our proposed method uses a random transformation to shuffle the pixels in the block. The $k^{th}$ pixel value is denoted by $p\left(k\right)$ in each block, where $1\le k\le s\times s$. Let $\alpha$ be a random permutation vector of the integers from 1 to $s\times s$. The new $k^{th}$ pixel value $p^\prime\left(k\right)$ after shuffling is given by $p^\prime\left(k\right)=p\left(\alpha_k\right)$. Fig. \ref{fig:props} illustrates the process of pixel shuffling transformation. 

This transformation is randomly applied to each image block in image $I$, and the permutation vector $\alpha$ is regenerated each time. Then we can get a shuffled image $I_P$. We can only shuffle part of blocks instead of all by adjusting the ratio parameter $q$. The reason why we regenerate the vector $\alpha$ every time the blocks are shuffled is that we find that if the same vector $\alpha$ is used for shuffling transformation, the neural network can restore the local spatial information destroyed by a single transformation operation through learning. But our goal is to occlude local areas by shuffling, so we discourage the network from restoring spatial information.

For deepfake detection tasks, the neural network predicts whether the input image $I$ is real or fake. Thus we label each image using a scalar $y$ as 1 if the input image is fake, otherwise it is 0. The classification network $C$ maps the input image $I$ to a probability variable $C_\Theta\left(I\right)$, where $\Theta$ are learnable parameters in the neural network. Using the shuffled image $I_P$ for training, the cross-entropy loss $\ell_{cls}\left(\Theta\right)$ can be written as:
\begin{equation}
  \ell_{cls}\left(\Theta\right)=-\sum{y\bullet\log{\left[C_\Theta\left(I_P\right)\right]}}
  \label{eq:eq1}
\end{equation}

\subsubsection{\color{black}Adversarial Loss}

\begin{figure}[htb]
  \centering
   \includegraphics[width=\linewidth]{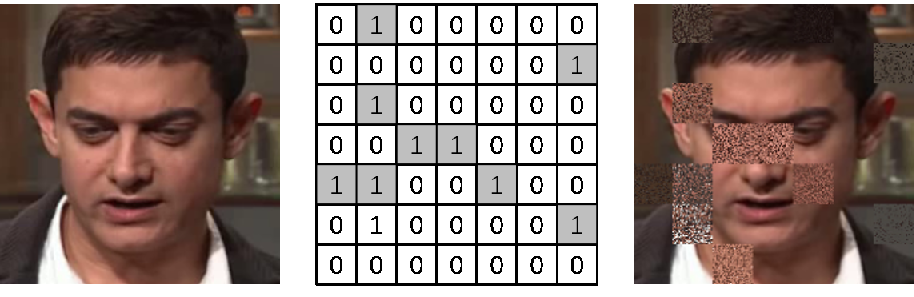}

   \caption{Examples for original images (left), intra-block shuffled images (right) and marked matrixs (middle).}
   \label{fig:examps}
\end{figure}

Neural networks also describe random noise and error while learning the underlying data distribution~\cite{Zhang2021UnderstandingDL}. As shown in Fig. \ref{fig:examps}, the shuffling transformation introduces noise patterns. The features learned by the neural network from these noise patterns mislead the detection task, which increases the false positive rate in the results. Adversarial loss is efficient for preserving domain-invariant patterns and rejecting domain-specific patterns~\cite{Tzeng2017AdversarialDD, Liu2019TransferableAT, Chen2019TransferabilityVD}. Thus, 
we propose an adversarial loss to reject patterns unrelated to deepfake detection to prevent the network from being affected by noise patterns introduced by shuffling.

\begin{figure}[htb]
  \centering
   \includegraphics[width=\linewidth]{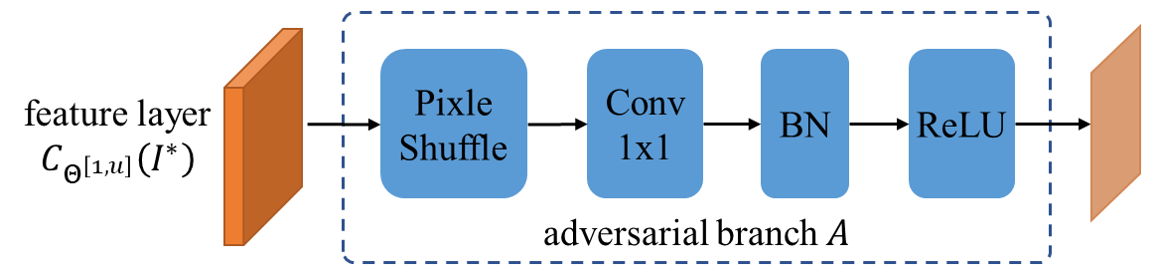}

   \caption{The structure of adversarial branch.}
   \label{fig:advbranch}
\end{figure}

Given image $I$ and its corresponding shuffled version $I_P$, we get a $m\times n$ matrix $P$. $P_{i,j}$ represents the element in the $i^{th}$ row and $j^{th}$ column corresponding to the block $B_{i,j}$ in the image $I_P$. $P_{i,j}$ is 1 if the image block $B_{i,j}$ is shuffled, otherwise it is 0. An adversarial network $A$ can be added as a new branch in the framework to output a matrix with a size of $m\times n$ to judge whether image blocks are shuffled or not. The intra-block adversarial loss $\ell_{adv}\left(\Theta^{\left[1,\ u\right]}, \Psi\right)$ can be computed as:
\begin{equation}
  \ell_{adv}\left(\Theta^{\left[1,\ u\right]}, \Psi\right)=-\sum{P\bullet\log{\left[A_\Psi\left(C_{\Theta^{\left[1,\ u\right]}}\left(I_P\right)\right)\right]}}
  \label{eq:eq2}
\end{equation}
% \begin{equation}
%   \tilde{P}\left(\varphi\left(I\right)\right)=A_\Psi\left(C_{\Theta^{\left[1,\ u\right]}}\left(\varphi\left(I\right)\right)\right)
%   \label{eq:eq2}
% \end{equation}
where $\Theta^{\left[1,\ u\right]}$ is the learnable parameters from the $1^{st}$ layer to $u^{th}$ layer in the neural network, $C_{\Theta^{\left[1,\ u\right]}}\left(I_P\right)$ is the output features of the $u^{th}$ layer in the neural network, $A$ represents a region alignment network work on the output features, and $\Psi$ is its parameters. As shown in Fig. \ref{fig:advbranch}, we use a PixleShuffle\cite{Shi2016RealTimeSI} instead of upscaling to align the input features with the mark matrix $P$. The features are processed by $1\times 1$ convolution to obtain outputs with one channel. Then the outputs are handled by a ReLU to get a matrix with the size of $m\times n$. 

% The intra-block adversarial loss $\ell_{adv}$ can be computed as:
% \begin{equation}
%   \ell_{adv}=-\sum{P\bullet\log{\left[\tilde{P}\left(\varphi\left(I\right)\right)\right]}}
%   \label{eq:eq3}
% \end{equation}

\subsection{Inter-block shuffling}
\label{sec:inter}

In deepfake detection tasks, the focus is primarily directed towards local features rather than global information. This emphasis is attributed to the fact that facial images tend to exhibit a consistent global structure, while the significant differences between real and fake facial images typically manifest in local regions. Therefore, the accurate identification of specific differences within local areas plays a important role in effective deepfake detection. Moreover, an undue emphasis on global information by neural networks may lead to a degradation of the deepfake detection process, converging it towards face recognition. To address this problem, we propose a novel approach that disrupts the global structure through inter-block shuffling, so that the network can focus on the extraction of features from local areas. By inducing changes in the locations of image patches, we establish a function that indirectly facilitates weight-sharing among local regions and enhances the translation invariance capabilities of the network. Additionally, we introduce a position restoration method to preserve the correlation of local features without requiring labels. By implementing these strategies, our proposed method aims to optimize deepfake detection by augmenting the focus on local features while mitigating the potential for misclassification stemming from an overemphasis on global information.

% For deepfake detection tasks, we pay more attention to local features than global information. Because face images always have a similar global structure, the significant differences between real and fake images usually appear in local areas. The specific differences in local areas of the image are essential bases for detection. In addition, the attention of neural networks on global information may make deepfake detection degenerate into face recognition. We propose to destroy the global structure through inter-block shuffling and encourage the network to extract features from local areas. Changes in image patch locations indirectly result in weight-sharing between local areas and enhance the translation invariance capabilities of the network. Meanwhile, a position restoration method is proposed to maintain the correlation of local features without labels. 

\subsubsection{Block shuffling transformation}

\begin{algorithm}[htb]
  \caption{Block Shuffling Procedure.}
  \label{alg:alg1}
  \begin{algorithmic}
  \STATE 
  \STATE \textbf{Input:} Input image $I$; Block size $s$; The numbers of horizontal blocks $m$; The numbers of horizontal blocks $n$; Blocks of image $B$; intra-shuffling ratio $q_a$; inter-shuffling ratio $q_b$.
  \STATE \textbf{Output:} Shuffled image $I^*$; Matrix $P$ marks the intra-shuffling of the blocks; Matrix $M$ marks the inter-shuffling of the blocks.
  \STATE $q_1 \leftarrow Rand(0, 1)$;
  \IF {$q_1 > q_b$}
    \STATE generate the matrix $M$;
    \STATE inter-shuffle image $I$;
  \ENDIF
  \FOR {$i \leftarrow 1$ to $m$}
    \FOR {$j \leftarrow 1$ to $n$}
      \STATE $q_2 \leftarrow Rand(0, 1)$;
      \IF {$q_2 > q_a$}
        \STATE intra-shuffle block $B_{i,j}$;
        \STATE $P_{i,j} \leftarrow 1$;
      \ELSE
        \STATE $P_{i,j} \leftarrow 0$;
      \ENDIF
    \ENDFOR
  \ENDFOR
  \end{algorithmic}
  \label{alg1}
\end{algorithm}

\begin{figure}[htb]
  \centering
   \includegraphics[width=0.8\linewidth]{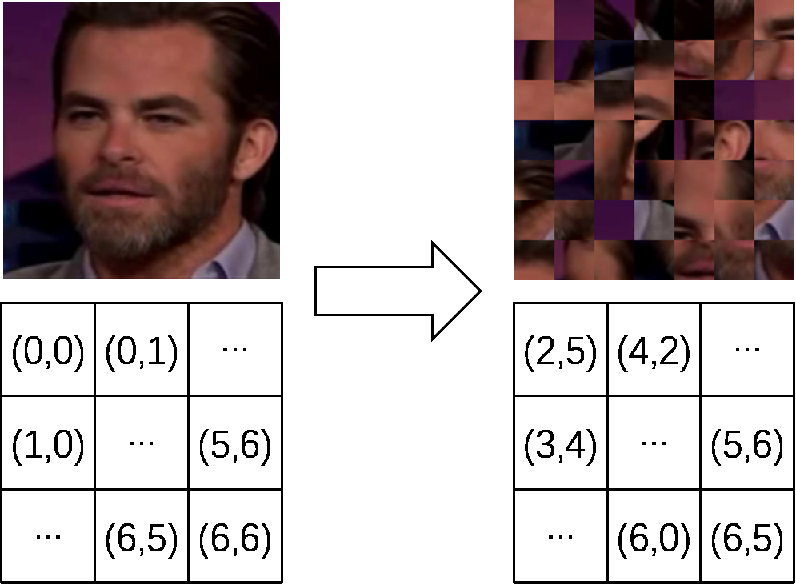}

   \caption{Example images for original images (left), inter-block shuffled images (right) and the coordinates of image blocks (bottom).}
   \label{fig:exambs}
\end{figure}

In Sec. \ref{sec:ps} we have divided image $I$ into blocks with a size of $m\times n$. As shown in Fig. \ref{fig:exambs}, we use a similar method to arrange and shuffle the image blocks to obtain a new image block arrangement. Then we can get a shuffled image $I_B$. The new image block $B_{i,j}^\prime$ after shuffling is given by $B_{i,j}^\prime=B_{\beta\left(i,j\right)}$, where $\beta\left(i,j\right)$ represents the original coordinates in image $I$ before being shuffled. Using two shuffling methods for training, the shuffled algorithm flow is shown in Algorithm \ref{alg:alg1}. To conveniently obtain the label matrix $P$ of intra-block shuffling, we first perform inter-block shuffling and then intra-block shuffling to get the shuffled image $I^*$. The cross-entropy loss $\ell_{cls}\left(\Theta\right)$ can be written as:
\begin{equation}
  \ell_{cls}\left(\Theta\right)=-\sum{y\bullet\log{\left[C_\Theta\left(I^*\right)\right]}}
  \label{eq:eq4}
\end{equation}

\subsubsection{Block position restoration}

Some discernible signs of forgery consistently emerge in specific regions of manipulated images, such as inconsistencies in eye color, the absence of nostril shadows, or misalignments in teeth structures. These localized imperfections represent important cues that neural networks utilize to identify fake images. Therefore, we propose the implementation of a position restoration algorithm to preserve the correlation of these distinctive attributes\cite{Noroozi2016UnsupervisedLO, Chen2019DestructionAC}. For the human observer, even when presented with only a small portion of a facial image, it is often possible to ascertain its origin within the face, such as the eye or nose region, and subsequently infer its approximate position within the complete facial image. In a similar vein, our objective is to enable the neural network to acquire this intuitive understanding and extract more effective features for the detection of deepfake images. By this approach, we aim to imbue the neural network with a human-like understanding, allowing it to discern and localize significant facial components, thus enhancing its ability to detect deepfake images more accurately.

% Partial forgery traces always appear in specific areas of the forged image, such as different eye colours, no shadows on the nostrils, and misalignment of teeth in the mouth. These blemishes that only appear in specific areas of the images are an important basis for neural networks to detect forged images. Thus, we propose to use a position restoration algorithm to preserve this correlation\cite{Noroozi2016UnsupervisedLO, Chen2019DestructionAC}. For the human eye, even if there is only a small part of the face image, we can easily determine which part of the face (such as the eye or nose) it comes from, and then get the approximate position in the face image. We aim to let the neural network learn this knowledge in the human way to extract better features to detect deepfake images.

\begin{figure}[htb]
  \centering
   \includegraphics[width=\linewidth]{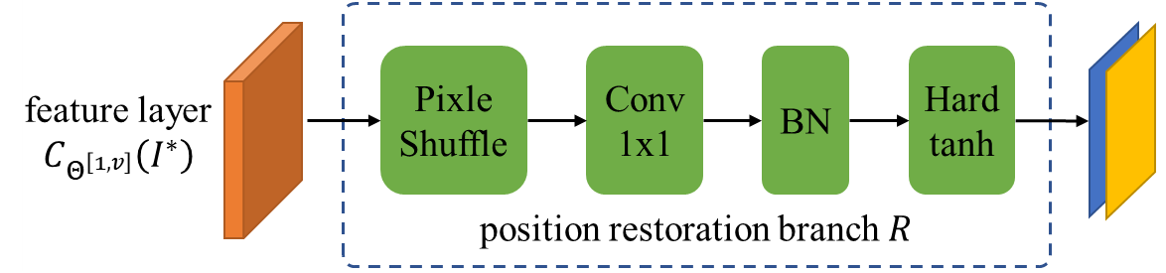}

   \caption{The structure of position restoration branch.}
   \label{fig:reconbranch}
\end{figure}

Given image $I$ and its corresponding inter-block shuffled version $I^*$, we get a $2\times m\times n$ matrix $M$. $M_{0,i,j}$ and $M_{1,i,j}$ are the row coordinates and column coordinates of the coordinates $\beta\left(i,j\right)$, respectively. An inter-block position restoration branch $R$ extracts the output features of the $v^{th}$ layer of the classification network and outputs the original coordinates corresponding to each image block. The inter-block restoration loss $\ell_{loc}\left(\Theta^{\left[1,v\right]}, \Phi\right)$ can be expressed as:
\begin{equation}
  \ell_{loc}\left(\Theta^{\left[1,v\right]}, \Phi\right)=-\sum\left|M-R_\Phi\left(C_{\Theta^{\left[1,v\right]}}\left(I^*\right)\right)\right|_1
  \label{eq:eq5}
\end{equation}
% \begin{equation}
%   R\left(\psi\left(I\right)\right)=R_\Phi\left(C_{\Theta^{\left[1,v\right]}}\left(\psi\left(I\right)\right)\right)
%   \label{eq:eq5}
% \end{equation}
where $\Phi$ are the parameters in the branch $R$. As shown in Fig. \ref{fig:reconbranch}, the branch $R$ consists of a PixleShuffle, a $1\times 1$ convolution and a Hardtanh to get a map with the size of $2\times m\times n$. 
% The inter-block reconstruction loss $\ell_{loc}$ can be expressed as:
% \begin{equation}
%   \ell_{loc}=-\sum\left|M-R\left(\psi\left(I\right)\right)\right|_1
%   \label{eq:eq6}
% \end{equation}

\subsection{Loss function}

Our framework completes the deepfake detection task through an end-to-end framework. As shown in Algorithm \ref{alg:alg2}, firstly, the input image $I$ enters the shuffling module to get a shuffled image $I^*$, a mark matrix $P$ records the intra-shuffling information, and a coordinate matrix $M$ records the inter-shuffling information. The shuffled image $I^*$ input classification network $C$ and get the result of forgery detection. The $u^th$ and $v^th$ feature layers of network $C$ are input to the adversarial branch $A$ and the restoration branch $R$, respectively. These branches help to suppress random noise and learn unique features based on the correlation of various blocks in the face image. Therefore, the neural network can extract a set of image features with good discrimination and generalization. Our method can work on most CNNs with almost no additional computational cost. Specifically, we want to minimize the following objective:
\begin{equation}
  \begin{split}
    \ell\left(\Theta,\Psi,\Phi\right)=\ell_{cls}\left(\Theta\right)+\alpha\ell_{adv}\left(\Theta^{\left[1,\ u\right]}, \Psi\right)\\
    +\beta\ell_{loc}\left(\Theta^{\left[1,v\right]}, \Phi\right)
    \label{eq:eq7}
  \end{split}
\end{equation}
where $\alpha$ and $\beta$ are the balancing hyperparameters.

\begin{algorithm}[htb]
  \caption{Block Shuffling learning framework Procedure.}
  \label{alg:alg2}
  \begin{algorithmic}
  \STATE 
  \STATE \textbf{Input:} Training set ${\{\left(I^{\left(n\right)},y^{\left(n\right)}\right)\}}^{N}_{n=1}$; the learning rate $\gamma$; the maximum number of iterations $T$.
  \STATE \textbf{Initialization:} $\mathbf{w} \leftarrow 0$; $\mathbf{u} \leftarrow 0$; $\mathbf{v} \leftarrow 0$; $t \leftarrow 0$.
  \REPEAT
  \FOR {$n \leftarrow 1$ to $N$}
  \STATE generate shuffled image $I^*$, matrix $P$ and $M$ based on Algorithm \ref{alg:alg1}
  \STATE $\mathbf{w} \leftarrow \mathbf{w} + y^{\left(n\right)} \bullet \log\left[C_\Theta\left(I^*\right)\right]$
  \STATE $\mathbf{u} \leftarrow \mathbf{u} + P \bullet \log{\left[A_\Psi\left(C_{\Theta^{\left[1,\ u\right]}}\left(I^*\right)\right)\right]}$
  \STATE $\mathbf{v} \leftarrow \mathbf{v} + \left|M - R_\Phi\left(C_{\Theta^{\left[1,v\right]}}\left(I^*\right)\right)\right|_1$
  \ENDFOR
  \STATE $\Theta \leftarrow \Theta - \gamma\nabla\left(\mathbf{w}+\alpha\mathbf{u}+\beta\mathbf{v}\right)$
  \STATE $\Psi \leftarrow \Psi - \gamma\nabla\alpha\mathbf{u}$
  \STATE $\Phi \leftarrow \Phi - \gamma\nabla\beta\mathbf{v}$
  \STATE $t \leftarrow t+1$
  \UNTIL $t=T$
  \end{algorithmic}
  \label{alg2}
\end{algorithm}

\section{Experiments}
\label{sec:exper}

In this section, we first introduce the experimental setups and then use a large number of experimental results to show the superiority of our method.

\subsection{Experimental settings}

{\bf Datasets}: We conduct experiments on FaceForensics++ (FF++)\footnote{https://github.com/ondyari/FaceForensics}~\cite{Rssler2019FaceForensicsLT}, Celeb-DF(v2)\footnote{https://github.com/yuezunli/celeb-deepfakeforensics}~\cite{Li2020CelebDFAL} and DeeperForensics-1.0 (DF1.0)\footnote{https://github.com/EndlessSora/DeeperForensics-1.0}~\cite{Jiang2020DeeperForensics10AL}. FF++ is a large-scale video dataset consisting of 1,000 original videos with real faces that have been manipulated by four deepfake methods~\cite{DeepFakes, Thies2019Face2FaceRF, FaceSwap, Thies2019DeferredNR}. \textcolor{black}{On this basis, FF++ has two subsets FaceShifter~\cite{Li2019FaceShifterTH} and Deep Fake Detection Dataset provided by Google \& JigSaw\footnote{https://ai.googleblog.com/2019/09/contributing-data-to-deepfake-detection.html}. The use of these two subsets in this paper is described separately.} Celeb-DF(v2) includes 590 original videos collected from YouTube and 5639 deepfake videos generated by an improved deepfake synthesis algorithm, which brings better visual quality. DF1.0 contains 60000 videos, where fake videos are generated by their DF-VAE method.
The dataset performs more challenging benchmarks with larger scale and higher diversity by applying a wide range of real-world perturbations on videos.

{\bf Evaluation Metrics}: We mainly apply the Accuracy score (ACC) and the Area Under the RoC Curve (AUC) as our evaluation metrics, which are commonly used in deepfake detection tasks~\cite{Zhou2017TwoStreamNN,Li2019ExposingDV,Wang2021M2TRMM,Du2020TowardsGD,Wang2021RepresentativeFM,Zhao2021MultiattentionalDD}.

{\bf Implementation details}: For all real and fake videos, we use a state-of-the-art efficient face detector BlazeFace~\cite{Bazarevsky2019BlazeFaceSN} to detect the faces of each frame and resize the facial images as inputs with a size of $224\times 224$. Random horizontal flip is applied for data augmentation. For intra-block shuffling, we set the block size $s$ to 16. We set the ratio parameter $q\sim U\left(0.4,0.6\right)$ of intra-block shuffling. For inter-block shuffling, we set the block size $s$ to 32. We apply the inter-block shuffling to all blocks of the image. We set the hyperparameters $\alpha=1$ and $\beta=1$ by default. We adopt Xception~\cite{Chollet2017XceptionDL} as the backbone network of our BSL framework. We use Adam~\cite{Kingma2015AdamAM} for optimization with a learning rate of $1e^{-4}$ and weight decay of $1e^{-6}$. We train our models on an Nvidia A10 GPU with a batch size of 64. 

\subsection{Evaluation of Intra-datasets}

Detecting fake images using the same forgery method as the training set is the basic ability of the deepfake detection model. FF++ is the most widely used dataset to test the basic capabilities of deepfake detection algorithms. FF++ contains multiple video quality versions. We train and test the HQ (c23) and the LQ (c40) versions. As the results shown in Table \ref{tab:tab7}, our method achieves state-of-the-art performance on the HQ version. However, since M2TR~\cite{Wang2021M2TRMM} incorporates a multi-model and multi-scale feature extractor for the generalization of image compression, our method performs 0.18\% lower than M2TR in the LQ version. In contrast, our method achieves similar results with almost no additional computation. Moreover, our method does not conflict with M2TR and can work together with the network. \textcolor{black}{As shown in Table \ref{tab:tab9}, our proposed method achieves good performance on all experimental datasets, which demonstrates the effectiveness of the method in the face of various forgery methods.}

\begin{table}[htb]
  \centering
  \caption{Quantitative frame-level detection results (AUC (\%)) on FF++ dataset with the LQ version and the HQ version, respectively. Results of CFFs are cited directly from \cite{Yu2022IGCL} and other methods are cited directly from \cite{Zhao2021MultiattentionalDD}. The best results are marked as bold.}
  \label{tab:tab7}
  \begin{tabular}{l|c|c} 
  \hline
  \makebox[0.15\textwidth][l]{Methods}  & \makebox[0.05\textwidth][c]{LQ}  & \makebox[0.05\textwidth][c]{HQ}  \\ 
  \hline
  % MesoNet\cite{Afchar2018MesoNetAC}                 & 70.47          & -              & 83.10          & -               \\
  Xception\tablefootnote{https://github.com/tstandley/Xception-PyTorch}~\cite{Chollet2017XceptionDL}                & 89.30          & 96.30           \\
  DSP-FWA\tablefootnote{https://github.com/yuezunli/DSP-FWA}~\cite{Li2019ExposingDV}                 & 92.49             & 99.28           \\
  % Two Branch\tablefootnote{https://github.com/yuezunli/DSP-FWA}~\cite{Masi2020TwoBranchRN}              & -              & 86.59          & -              & 98.70           \\
  M2TR\tablefootnote{https://github.com/wangjk666/M2TR-Multi-modal-Multi-scale-Transformers-for-Deepfake-Detection}~\cite{Wang2021M2TRMM}                    & \textbf{94.25} & 99.05           \\
  SRM\tablefootnote{https://github.com/crywang/face-forgery-detection}~\cite{Luo2021GeneralizingFF}                     & 91.36              & 98.70           \\
  Multi-att\tablefootnote{https://github.com/yoctta/multiple-attention}~\cite{Zhao2021MultiattentionalDD}               & 87.26          & 98.97           \\ 
  CFFs\tablefootnote{https://github.com/botianzhe/CFFExtractor}~\cite{Yu2022IGCL}               & 90.35          & 97.63           \\
  \hline
  Ours(Xception)                    & 93.13          & \textbf{99.95}  \\
  \hline
  \end{tabular}
\end{table}

\begin{table}[htb]
  \centering
  \caption{Intra-datasets results(AUC (\%)) on six subsets of FF++, Celeb-DF(v2) and DF1.0}
  \label{tab:tab9}
  \begin{tabular}{l|c|c} 
  \hline
  \makebox[0.15\textwidth][l]{Methods}  & \makebox[0.05\textwidth][c]{ACC}  & \makebox[0.05\textwidth][c]{AUC}  \\ 
  \hline
  % MesoNet~\cite{Afchar2018MesoNetAC}        & 76.2  & -      \\
  
  Deepfakes        & 99.09 & 99.95  \\ 
  Face2Face        & 99.90 & 99.99  \\ 
  FaceSwap        & 99.19 & 99.89  \\ 
  NeuralTextures       & 95.63 & 97.59  \\ 
  FaceShifter       & 98.12 & 99.54  \\ 
  Google \& JigSaw       & 95.15 & 98.64  \\ 
  Celeb-DF       & 98.26 & 99.72  \\
  DF1.0            & 99.99 & 99.99  \\
  \hline
  \end{tabular}
\end{table}

\color{black}
\subsection{Cross-dataset evaluation}

It is possible for a deepfake detection model to encounter fake methods that it did not encounter during the training phase. To evaluate the performance of the detection method in this case, we trained the model on the HQ version of FF++ and tested the model on Celeb-DF(v2). The results are shown in Table \ref{tab:tab8}. Our method outperforms most of the existing methods, especially better than our model's backbone network Xception. Although Two Branch~\cite{Masi2020TwoBranchRN} is slightly higher than ours in cross-dataset performance, its performance on the original dataset FF++ is far inferior to ours.

\begin{table}[htb]
  \centering
  \caption{Cross-dataset evaluation(AUC(\%)) on Celeb-DF(v2) 
  by training on the HQ version of FF++. Results of some other methods are
  cited directly from ~\cite{Zhao2021MultiattentionalDD}. The best results are marked as bold.}
  \label{tab:tab8}
  \begin{tabular}{l|c|c} 
  \hline
  \makebox[0.15\textwidth][l]{Methods}  & \makebox[0.08\textwidth][c]{FF++}  & \makebox[0.08\textwidth][c]{Celeb-DF(v2)}  \\ 
  \hline
  Two-stream~\cite{Zhou2017TwoStreamNN}        & 70.10  & 53.80      \\
  MesoNet\tablefootnote{https://github.com/HongguLiu/MesoNet-Pytorch}~\cite{Afchar2018MesoNetAC}        & 84.70  & 54.80      \\
  Mesolnception4~\cite{Afchar2018MesoNetAC}       & 83.00  & 53.60      \\
  FWA~\cite{Li2019ExposingDV}        & 80.10 & 56.90  \\ 
  Xception-raw~\cite{Chollet2017XceptionDL}       & 99.70 & 48.20  \\
  Xception-c23~\cite{Chollet2017XceptionDL}       & 99.70 & 65.30  \\
  Xception-c40~\cite{Chollet2017XceptionDL}       & 95.50 & 65.50  \\
  Capsule\tablefootnote{https://github.com/nii-yamagishilab/Capsule-Forensics}~\cite{Nguyen2019CapsuleforensicsUC}        & 96.60 & 57.50  \\ 
  DSP-FWA~\cite{Li2019ExposingDV}        & 93.00 & 64.60  \\ 
  Two Branch~\cite{Masi2020TwoBranchRN}        & 93.18 & \textbf{73.41}  \\ 
  Multi-att~\cite{Zhao2021MultiattentionalDD}    & \textbf{99.80} & 67.44 \\
  \hline
  Ours(Xception) & 99.72 & 69.10 \\
  \hline
  \end{tabular}
\end{table}

In order to more intuitively demonstrate the generalization of our method under different datasets, we train the model on each of the four methods on the HQ version of FF++ dataset and evaluate its performance on CelebDF(v2) and DF1.0. This experimental setting is very challenging. Each model is trained with only one type of fake sample, and the real samples and fake methods in the test sets are not repeated with the training sets. The results are shown in Table \ref{tab:tab10}. Our method achieves the best results in most experiments. Among them, Face X-ray~\cite{Li2020FaceXF} and SLADD~\cite{Chen2022SelfsupervisedLO} both add additional training data, and SRM~\cite{Luo2021GeneralizingFF} and F3Net~\cite{Qian2020ThinkingIF} rely on high-frequency components of images to detect deepfake. These settings can effectively improve generalization performance, so these methods have achieved better results than ours in some experiments. But our proposed method can also be combined with the above methods to achieve better performance.

\begin{table*}[htb]
  \centering
  \caption{Cross-dataset evaluation(AUC(\%)) on CelebDF(v2) and DF1.0 by training on each of the four methods on the HQ version of FF++. The first row denotes the training data, and the second row denotes the corresponding test data. Results of some other methods are
  cited directly from ~\cite{Chen2022SelfsupervisedLO}. The best results are marked as bold.}
  \label{tab:tab10}
  \begin{tabular}{l|c|c|c|c|c|c|c|c} 
  \hline
  \multirow{2}{*}{\makebox[0.15\textwidth][l]{Methods}} & \multicolumn{2}{c|}{DF}  & \multicolumn{2}{c|}{F2F}  & \multicolumn{2}{c|}{FS}  & \multicolumn{2}{c}{NT}  \\ 
  \cline{2-9}  & \makebox[0.05\textwidth][c]{CelebDF}  & \makebox[0.05\textwidth][c]{DF1.0}  & \makebox[0.05\textwidth][c]{CelebDF}  & \makebox[0.05\textwidth][c]{DF1.0}  & \makebox[0.05\textwidth][c]{CelebDF}  & \makebox[0.05\textwidth][c]{DF1.0}  & \makebox[0.05\textwidth][c]{CelebDF}  & \makebox[0.05\textwidth][c]{DF1.0}  \\ 
  \hline
  Xception~\cite{Chollet2017XceptionDL}  & 68.1  & 61.7  & 59.8  & 74.5  & 60.1  & 60.5  & 62.5  & 83.8  \\
  Face X-ray\tablefootnote{https://github.com/Daisy-Zhang/Face-X-ray}~\cite{Li2020FaceXF}  & 55.4  & 66.8  & 68.4  & 76.6  & 69.7  & \textbf{79.5}  & 70.3  & 86.6  \\
  F3Net\tablefootnote{https://github.com/Leminhbinh0209/F3Net}~\cite{Qian2020ThinkingIF}  & 66.4  & 65.8  & 65.4  & 76.1  & 63.6  & 65.1  & 68.9  & \textbf{93.2}  \\
  RFM\tablefootnote{https://github.com/crywang/RFM}~\cite{Wang2021RepresentativeFM}  & 72.3  & 71.7  & 66.3  & 73.2  & 59.1  & 71.4  & 60.0  & 84.6  \\
  SRM~\cite{Luo2021GeneralizingFF}  & 65.0  & 72.0  & 69.3  & 77.5  & 64.3  & 77.1  & 65.1  & 93.6  \\
  SLADD\tablefootnote{https://github.com/liangchen527/SLADD}~\cite{Chen2022SelfsupervisedLO}  & 73.0  & 74.2  & \textbf{78.1}  & 78.6  & \textbf{80.0}  & 69.5  & \textbf{75.9}  & 88.9  \\
  \hline
  Ours(Xception)  & \textbf{73.7}  & \textbf{82.0}  & 67.1  & \textbf{92.7}  & 62.6  & 76.9  & 69.8  & 87.0  \\
  \hline
\end{tabular}
\end{table*}
\color{black}

\color{black}
\subsection{Cross video quality evaluation}

The quality of the video will be affected when it spreads in the network, and this quality degradation may affect the performance of the deepfake detection model. FF++ provides video versions compressed under H264 video encoding by ffmpeg software. The HQ version and the LQ version correspond to the c23 and c40 compression settings respectively. We simulate the video quality degradation experienced by deepfake detection models by training the HQ version and testing on the LQ version. The results are listed in Table \ref{tab:tab2}. We reproduce three existing methods and compare with our method. Although the results illustrate that the metrics of all models have dropped significantly and are lower than the results trained with the LQ version. Our method shows better generalization than SRM ~\cite{Luo2021GeneralizingFF} using frequency domain information and DSP-FWA~\cite{Li2019ExposingDV} using multi-scale design in the case of video compression.

\begin{table}[htb]
  \centering
  \caption{ Cross video quality evaluation(ACC(\%) and AUC(\%)) on the LQ version of FF++ by training on the HQ version. The best results are marked as bold.}
  \label{tab:tab2}
  \begin{tabular}{l|c|c} 
  \hline
  \makebox[0.15\textwidth][l]{Methods}  & \makebox[0.05\textwidth][c]{ACC}  & \makebox[0.05\textwidth][c]{AUC}  \\ 
  \hline
  Xception~\cite{Chollet2017XceptionDL}       & 83.45  & 93.13  \\
  SRM~\cite{Luo2021GeneralizingFF}     & 84.05  & 91.33  \\
  DSP-FWA~\cite{Li2019ExposingDV}      & 85.98  & 92.84  \\ 
  \hline
  Ours(Xception) & \textbf{87.41}  & \textbf{95.89}  \\
  \hline
  \end{tabular}
\end{table}
\color{black}

\begin{table*}[htb]
  \centering
  \caption{ Cross video quality evaluation(ACC(\%) and AUC(\%)) on the LQ version of FF++ by training on the HQ version. The best results are marked as bold.}
  \label{tab:tab2}
  \begin{tabular}{l|c|c|c|c|c|c|c} 
  \hline
  \makebox[0.1\textwidth][l]{Methods}  & \makebox[0.05\textwidth][c]{Attack}  & \makebox[0.05\textwidth][c]{Random} & \makebox[0.05\textwidth][c]{PD} & \makebox[0.05\textwidth][c]{WD+SR} & \makebox[0.05\textwidth][c]{ComDefend} & \makebox[0.05\textwidth][c]{Con} & \makebox[0.05\textwidth][c]{Ours} \\ 
  \hline
  Clean    & 100  & 99.68  & 83.50 & 86.34 & 93.80 & 98.74 & 97.70 \\
  FGSM     & 33.84  & 53.52  & 53.86 & 60.98 & 56.53 & 97.90 & 97.52 \\
  BIM      & 0.06  & 17.72  & 60.76 & 69.22 & 44.15 & 92.00 & 95.35 \\
  C\&W      & 1.90  & 93.32  & 84.81 & 87.50 & 90.39 & 96.50 & 97.00 \\
  DeepFool  & 0.68  & 90.94  & 84.32 & 87.36 & 89.09 & 96.48 & 96.21 \\
  PGD      & 0.02  & 18.50  & 66.44 & 79.42 & 50.91 & 93.14 & 94.82 \\
  MI-FGSM      & 0.04  & 8.96  & 39.44 & 65.08 & 34.17 & 87.00 & 90.46 \\
  DI-FGSM      & 1.30  & 2.80  & 60.68 & 69.00 & 41.77 & 65.44 & 70.34 \\
  MD-FGSM      & 1.18  & 2.28  & 21.10 & 29.62 & 20.53 & 75.26 & 81.26 \\
  APGD-CE      & 0.00  & 18.14 & 69.13 & 70.68 & 41.40 & 96.94 & 97.18 \\
  APGD-DLR     & 0.00  & 66.38  & 77.47 & 77.64 & 45.48 & 96.78 & 97.02 \\
  \hline
  \end{tabular}
\end{table*}
\color{black}

\color{black}
\subsection{Resizing and blurring evaluation}
\begin{figure}[htb]
  \centering
   \includegraphics[width=1.0\linewidth]{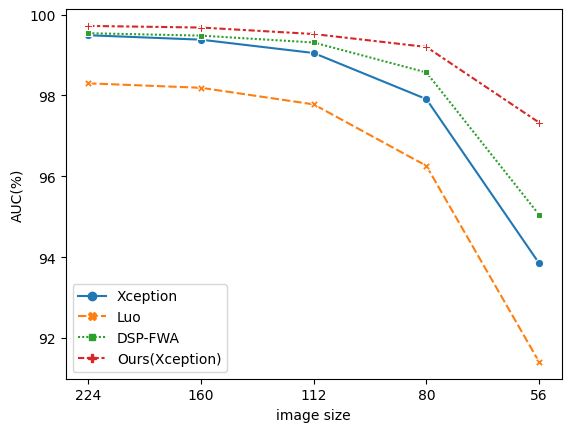}

   \caption{Resizing evaluation (AUC(\%)) on Celeb-DF v2 dataset with five different input image sizes.}
   \label{fig:resizeeff}
\end{figure}

\begin{figure}[htb]
  \centering
   \includegraphics[width=1.0\linewidth]{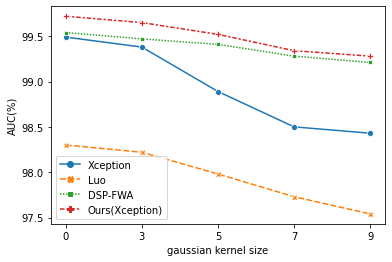}

   \caption{Blurring evaluation (AUC(\%)) on Celeb-DF v2 dataset with five different Gaussian blur kernel sizes.}
   \label{fig:blureff}
\end{figure}

We evaluate the performance of the forgery detection method under image transformations such as resizing and blurring, which are the most common transformations that images spread across the Internet. We chose Celeb-DF(v2) as the experimental data set, because the high-quality images provided by Celeb-DF(v2) are more suitable for testing the effect of the forgery detection model after the image quality is compromised. We reproduce three existing methods and compare with our method. We train models on Celeb-DF(v2) with an input image size of 224. Fig. \ref{fig:resizeeff} shows the AUC score compared with current methods under various input image sizes of 160, 112, 80 and 56, respectively. Fig. \ref{fig:blureff} shows the AUC score under various Gaussian kernel sizes of 3, 5, 7 and 9, respectively. Our method has achieved the best results as a whole, and as the intensity of transformation increases, we achieve more accurate results than other methods. DSP-FWA~\cite{Li2019ExposingDV} uses a dual spatial pyramid strategy on both image and feature level to solve multi-scale problems, so it achieves good results in this experiments, while our method uses a simple network structure to achieve slightly higher results.

\color{black}

%-------------------------------------------------------------------------
\section{Ablation Study}
\label{sec:abla}

\subsection{Effectiveness of block shuffling}

To evaluate the effectiveness of different modules, we combine them and use evaluations, respectively. All models are trained on the HQ version of FF++ dataset and tested on the HQ version and the LQ version of FF++ dataset. This setting separately evaluates the basic detection ability and generalization ability in the face of video quality degradation. The results are listed in Table \ref{tab:tab5}. 

Compared with model 1 (baseline Xception), model 2 (Xception with Intra-block shuffling) achieves a 1.69\% improvement on the HQ version but a 0.63\% drop on the LQ version. This is because although intra-block shuffling enhances detection, the introduced noise patterns limit the generalization ability. Model 3 (addition of adversarial learning) dramatically enhances performance and surpasses model 1 because it overcomes the overfitting caused by noise patterns due to the adversarial loss. Model 4 (addition of the Inter-block shuffling) saw a 0.77\% drop in performance on the HQ version because relying solely on the features extracted within the block significantly limits the detection ability of the model. We enhance local features by position restoration to build correlations in local regions. Model 5 (addition of the Block position restoration) achieves the best performance, and the ACC score increases by 2.91\% on the HQ version and 3.96\% on the LQ version. 

\begin{table}[htb]
  \centering
  \caption{ Ablation study of the proposed BSL. We compare BSL by gradually removing each module.}
  \label{tab:tab5}
  \begin{tabular}{c|c|c|c|c|c|c} 
  \hline
  \makebox[0.02\textwidth][c]{ID}  & \makebox[0.04\textwidth][c]{Intra} & \makebox[0.04\textwidth][c]{Adv} & \makebox[0.04\textwidth][c]{Inter} & \makebox[0.04\textwidth][c]{Restore} & \makebox[0.05\textwidth][c]{HQ}  & \makebox[0.05\textwidth][c]{LQ}  \\ 
  \hline
  1  &       &     &       &       & 95.73 & 83.45  \\
  2  & \checkmark      &     &       &       & 97.42 & 82.82  \\
  3  & \checkmark      & \checkmark    &       &       & 97.69 & 85.29  \\
  4  & \checkmark      & \checkmark    & \checkmark      &       & 96.92      & 85.64       \\
  5  & \checkmark      & \checkmark    & \checkmark      & \checkmark      & \textbf{98.64} & \textbf{87.41}  \\
  \hline
  \end{tabular}
\end{table}

\begin{figure}[htb]
  \centering
   \includegraphics[width=\linewidth]{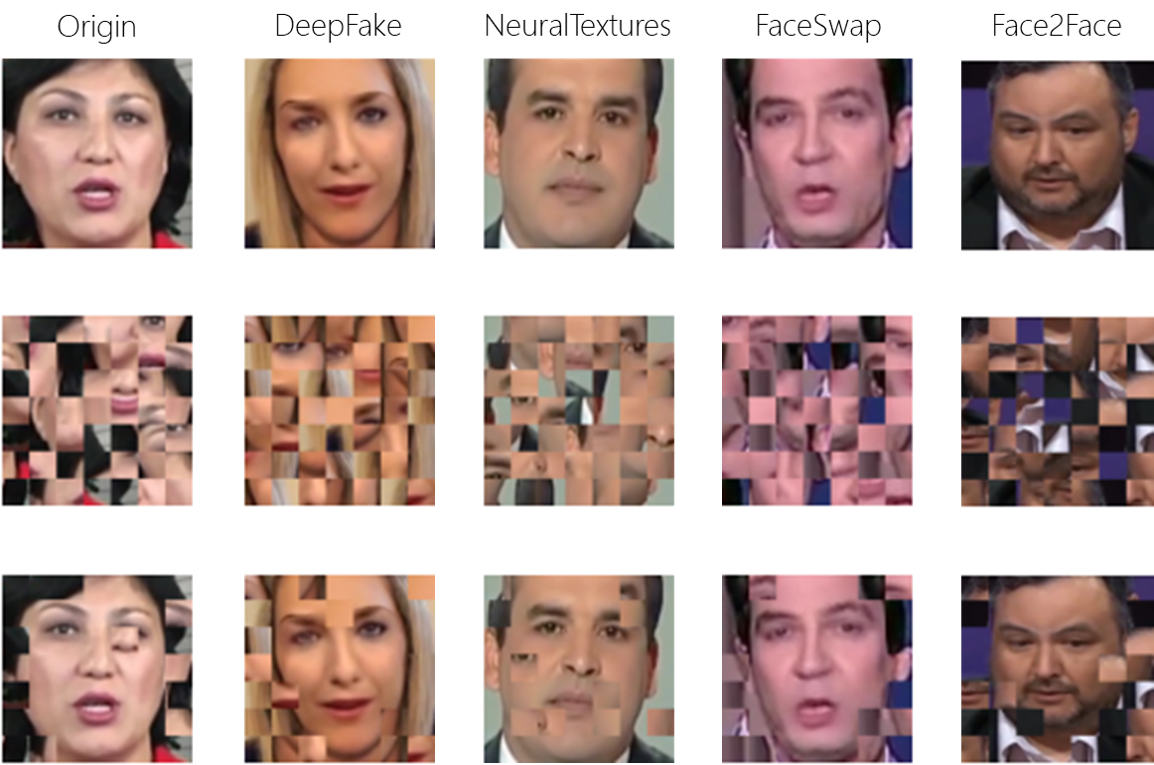}

   \caption{Examples for intact images (top), inter-block shuffling images (middle) and restore images (bottom) from ff++ dataset.}
   \label{fig:reconeff}
\end{figure}

\subsection{Effectiveness of block position restoration}

\begin{figure}[htb]
  \centering
   \includegraphics[width=1.0\linewidth]{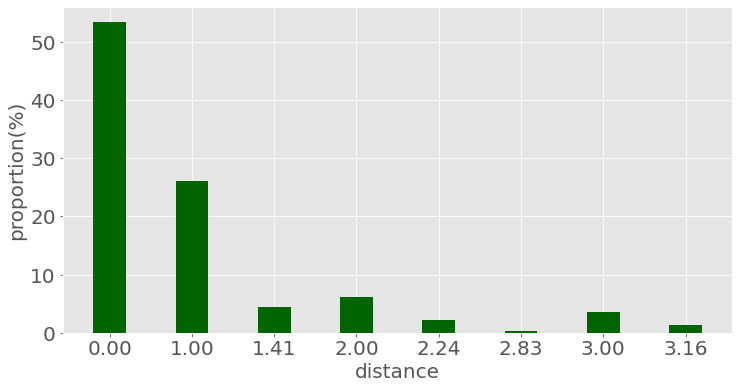}

   \caption{Evaluation of Block position restoration on the FF++ dataset. The distance between blocks before and after restoration is counted.}
   \label{fig:effRe}
\end{figure}

The Block position restoration is designed to help the network model the correlations between local features. The distance between blocks before and after restoration represents the effect of correlation modelling. We count the restoration results of the model trained on the HQ version of FF++ dataset. As shown in the Fig. \ref{fig:effRe}, the experimental results show that the distance before and after restoration of more than 80\% of blocks is less than or equal to 1. This means that most blocks are back in their original positions after restoration. Fig. \ref{fig:reconeff} intuitively illustrates the effect of inter-block shuffling and restoration.

\subsection{Effectiveness of block size for intra-block shuffling}

\begin{figure}[htb]
  \centering
   \includegraphics[width=1.0\linewidth]{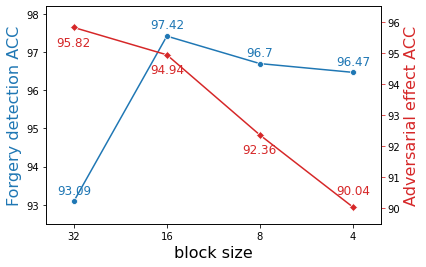}

   \caption{Evaluation of different block sizes for intra-block shuffling on the FF++ dataset with the HQ version. The blue line shows performance metrics for forgery detection, the red line shows performance metrics for adversarial effects.}
   \label{fig:bsintra}
\end{figure}

 To evaluate the impact of different block sizes, we count the forgery detection results and the results of the adversarial branch at different sizes on the HQ version of the ff dataset. Since the adversarial branch predicts whether each block is shuffled, we adopt ACC to measure how well the adversarial branch works. The results are shown in Fig. \ref{fig:bsintra}. For fake detection results, the detection performance is the best when the block size equals 16. But compared with other block sizes, the detection performance drops significantly when the block size equals 32. The choice of block size for intra-block shuffling is subtle. Too large a block causes intra-block shuffles to behave like global shuffles, which is disastrous for tasks such as forgery detection that value local regions. For the adversarial branch, the adversarial effect decreases slightly as the block size decreases. Because the smaller the blocks are, the more likely they are completely in the solid colour area (e.g. hair, forehead), and it is difficult to judge whether these blocks are shuffled or not.

\subsection{Effectiveness of block size for inter-block shuffling}

The block size of inter-block shuffling directly affects the granularity of local features and the difficulty of position restoration. The smaller block size, the less information it contains, and it is more challenging to build correlations between blocks. We count the forgery detection and restoration results of the models at different block sizes on the HQ version of the ff dataset. We use ACC to evaluate the detection performance and use the average distance difference before and after restoration to evaluate the restoration performance. To intuitively represent the restoration performance at different block sizes, we divide the distance difference by the block size to get the relative distance difference. The smaller the value, the better the restoration. 

\begin{figure}[htb]
  \centering
   \includegraphics[width=1.0\linewidth]{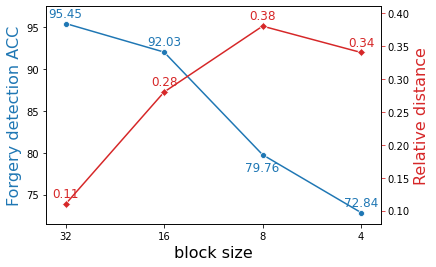}

   \caption{Evaluation of different block sizes for inter-block shuffling on the FF++ dataset with the HQ version. The blue line shows performance metrics for forgery detection, the red line shows performance metrics for restoration effects.}

   \label{fig:bsinter}
\end{figure}

As shown in Fig. \ref{fig:bsinter}, the experimental results show that the detection and reduction performance are the best when the block size equals 32. The detection performance and reduction performance decrease significantly as the block size decreases. Because the small block contains very little high-level feature information, almost only low-level colour information can be extracted, which significantly increases the difficulty of restoration. Meanwhile, the result of restoration affects the detection performance, and the detection performance also decreases.

\subsection{Effectiveness of hyperparameters}

\begin{figure}[htb]
  \centering
   \includegraphics[width=1.0\linewidth]{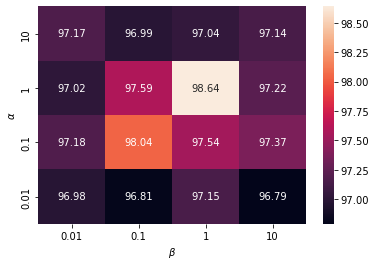}

   \caption{Evaluation (ACC) on the HQ version of FF++ dataset with different hyperparameters.}
   \label{fig:hyc23}
\end{figure}

\begin{figure}[htb]
  \centering
   \includegraphics[width=1.0\linewidth]{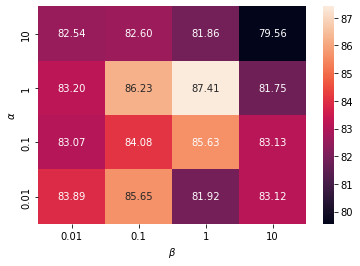}

   \caption{Cross video quality evaluation (ACC) on the LQ version of FF++ dataset by training on the HQ version with different hyperparameters.}
   \label{fig:hyc40}
\end{figure}

To evaluate the impact of hyperparameters $\alpha$ and $\beta$, we train on the HQ version of the ff++ dataset and test the HQ version and the LQ version under different parameter values. Fig. \ref{fig:hyc23} and Fig. \ref{fig:hyc40} show the ACC results tested on the HQ and LQ version datasets, respectively, and the best results are achieved when $\alpha=1, \beta=1$. When $\alpha$ and $\beta$ take large values, the centre of gravity of the network tends to adversarial learning and position restoration learning, two auxiliary tasks, so the forgery detection effect is not good. When $\alpha$ and $\beta$ take small values, the network degenerates into the backbone network, and the result of forgery detection is also close to the performance of the backbone network. When the hyperparameters take values 1 and 0.1, the model shows similar performance, with the highest performance when both hyperparameters are 1.

\subsection{Universality of BSL with various backbones}

In this section, we evaluate the versatility of our method. The above experiments are all based on Xception, and we also perform the same experiment on ResNet50\footnote{https://github.com/Lornatang/ResNet-PyTorch}~\cite{He2016DeepRL} and EfficientNet\footnote{https://github.com/lukemelas/EfficientNet-PyTorch}~\cite{Tan2019EfficientNetRM}. The results listed in Table \ref{tab:tab6} show that our proposed method plays an active role in various CNNs. The performance of the networks trained based on our method is significantly improved.

\begin{table}[htb]
  \centering
  \caption{ Quantitative frame-level detection results (AUC (\%)) on FF++ dataset. The best results are marked as bold.}
  \label{tab:tab6}
  \begin{tabular}{l|c|c} 
  \hline
  \makebox[0.15\textwidth][l]{Methods}  & \makebox[0.05\textwidth][c]{HQ}  & \makebox[0.05\textwidth][c]{LQ}  \\ 
  \hline
  ResNet50~\cite{He2016DeepRL}           & 96.04   & 80.90    \\
  Ours(ResNet50)     & \textbf{97.60}   & \textbf{86.79}    \\ 
  \hline
  EfficientNet-B4~\cite{Tan2019EfficientNetRM}       & 96.54   & 76.93    \\
  Ours(EfficientNet-B4) & \textbf{97.92}   & \textbf{85.61}    \\
  \hline
  \end{tabular}
\end{table}

%-------------------------------------------------------------------------
\section{Conclusion}
\label{sec:conc}

This paper discusses the significance of addressing the overfitting issue that arises within local regions during deepfake detection tasks, while also exploring its potential causes. Our focus has been on weight-sharing regularization as a means of investigating deepfake detection, with the introduction of the random shuffling method. To achieve weight-sharing across different dimensions indirectly, we propose a block shuffling learning framework that comprises an intra-block shuffling module and an inter-block shuffling module. The former module enhances network generalization by randomly occluding local areas and employs adversarial loss to suppress noise modes. On the other hand, the latter module aims to disrupt the global structure, encouraging the network to learn local features. Additionally, it employs position restoration algorithms to model the semantic correlation between local features. Our proposed method has demonstrated substantial improvements across a range of extensive metrics, as validated by comprehensive experimental results. It is important to note that our current approach primarily focuses on the spatial location of pixels within the image and does not involve alterations to pixel values. Moving forward, future work will consider augmenting network generalization by incorporating pixel-level random perturbations and exploring methods for reconstructing the original images during the training phase. These avenues of investigation hold promise for further enhancing the effectiveness of our proposed methodology.

% This paper discusses the importance and possible causes of the overfitting problem that occurs in local regions in the deepfake detection task. Then, we focus on weight-sharing regularization to research deepfake detection and introduce the random shuffling method. We propose a block shuffling learning framework consisting of an intra-block shuffling module and an inter-block shuffling module to achieve weight-sharing in different dimensions indirectly. Moreover, the former enhances the generalization of the network by randomly occluding the local areas and rejects the noise mode by using adversarial loss. The latter encourages the network to learn local features by destroying the global structure and models the semantic correlation between local features through position restoration algorithms. Our method has significantly improved in extensive metrics shown in extensive experimental results. Our method currently focuses on the location of pixels in the image and does not involve changes in pixel values. Therefore, in future work, we will consider improving the generalization of the network by adding pixel-level random perturbations and reconstructing the original images in the training phase.

% \bibliographystyle{IEEEtran}

% % \bibliography{egbib}

% \end{thebibliography}

\bibliographystyle{IEEEtran}
\bibliography{IEEEabrv, egbib}

\begin{IEEEbiography}[{\includegraphics[width=1in,height=1.25in,clip,keepaspectratio]{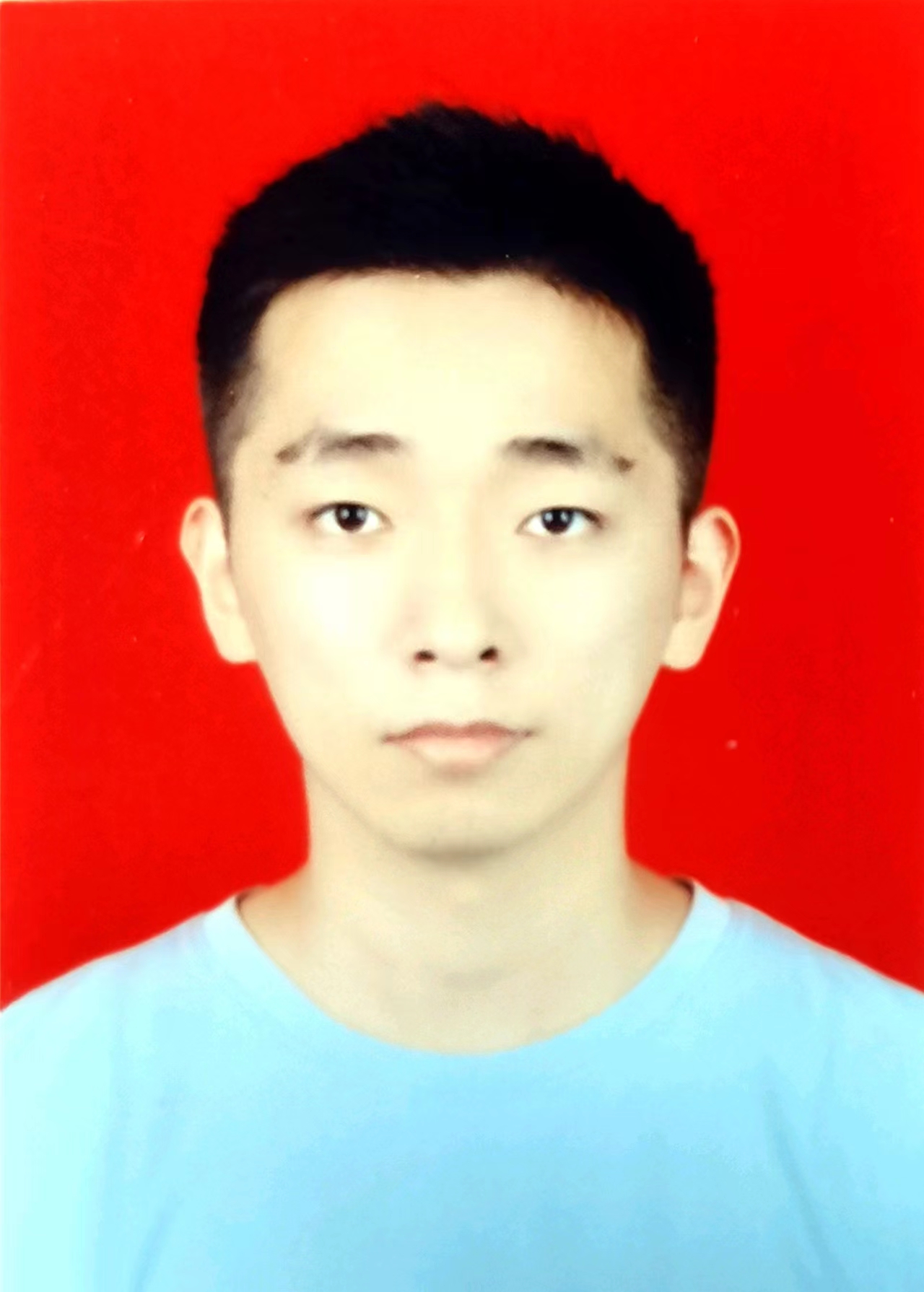}}]{Sitong Liu} received the B.S. degree in information and computing science from Nanjing University of Science and Technology in 2018, where he is currently pursuing the doctoral degree with the School of Computer Science. His reach interests include artificial intelligence security.
\end{IEEEbiography}

\begin{IEEEbiography}[{\includegraphics[width=1in,height=1.25in,clip,keepaspectratio]{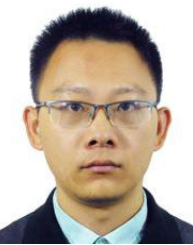}}]{Zhichao Lian} (Member, IEEE) received the Ph.D. degree in electrical and electronic engineering from Nanyang Technological University, Singapore, in 2013. From 2012 to 2014, he was a Postdoctoral Associate with the Department of Statistics, Yale University. He is currently an Associate Professor with the School of Computer Science and Engineering, Nanjing University of Science and Technology. His research interests include image processing, pattern recognition, and the IoT.
\end{IEEEbiography}


\begin{IEEEbiography}[{\includegraphics[width=1in,height=1.25in,clip,keepaspectratio]{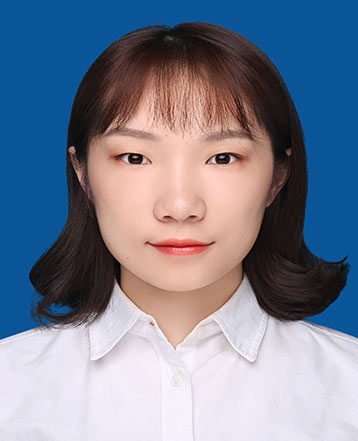}}]{Siqi Gu} received the B.S. degree in the School of Information Science and Technology, Beijing University of Chemical Technology in 2019 and the M.S. degree in computer science and engineering from Nanjing University of Science and Technology in 2022. Now she is currently working toward the doctoral degree in the School of Software Engineering from Nanjing University. Her research interests include intelligent software testing.
\end{IEEEbiography}

\begin{IEEEbiography}[{\includegraphics[width=1in,height=1.25in,clip,keepaspectratio]{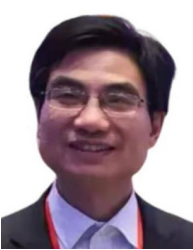}}]{Liang Xiao} (Member, IEEE) received the B.S. degree in applied mathematics and the Ph.D. degree in computer science from the Nanjing University of Science and Technology, Nanjing, China, in 1999 and 2004, respectively. \\
From 2009 to 2010, he was a Post-Doctoral Fellow at the Rensselaer Polytechnic Institute, Troy, NY, USA. He is currently a Professor with the School of Computer Science, Nanjing University of Science and Technology. His main research areas include inverse problems in image processing, scientific computing, data mining, and pattern recognition.
\end{IEEEbiography}

% \vfill

\end{document}